\documentclass[11pt,a4paper]{article}
\usepackage[hyperref]{acl2019}
\usepackage{times}
\usepackage{latexsym}

\usepackage{url}

\usepackage{graphicx}
\usepackage{amsmath}
\usepackage{amsfonts}
\usepackage{multirow}
\usepackage{longtable}
\newcommand{\tabincell}[2]{\begin{tabular}{@{}#1@{}}#2\end{tabular}}

\aclfinalcopy 
\def\aclpaperid{1766} 

\title{Lattice Transformer for Speech Translation}

\author{
  Pei Zhang\thanks{~~indicates equal contribution.}, Boxing Chen$^*$, Niyu Ge$^*$, Kai Fan$^*$\thanks{~~corresponding author.}\\
  Alibaba Group Inc.\\
  \texttt{\{xiaoyi.zp,boxing.cbx,niyu.ge,k.fan\}@alibaba-inc.com} \\}

\date{}

\begin{document}
\maketitle
\begin{abstract}
Recent advances in sequence modeling have highlighted the strengths of the transformer architecture, especially in achieving state-of-the-art machine translation results. 
However, depending on the up-stream systems, e.g., speech recognition, or word segmentation, the input to translation system can vary greatly.
The goal of this work is to extend the attention mechanism of the transformer to naturally consume the lattice in addition to the traditional sequential input. 
We first propose a general lattice transformer for speech translation where the input is the output of the automatic speech recognition (ASR) which contains multiple paths and posterior scores.
To leverage the extra information from the lattice structure, we develop a novel controllable lattice attention mechanism to obtain latent representations. 
On the LDC Spanish-English speech translation corpus, our experiments show that lattice transformer generalizes significantly better and outperforms both a transformer baseline and a lattice LSTM. 
Additionally, we validate our approach on the WMT 2017 Chinese-English translation task with lattice inputs from different BPE segmentations. In this task, we also observe the improvements over strong baselines.
\end{abstract}

\section{Introduction}

Transformer based encoder-decoder framework \cite{vaswani2017attention} for Neural Machine Translation (NMT) has currently become the state-of-the-art in many translation tasks, significantly improving translation quality in text \cite{bojar2018findings,fan2018bilingual} as well as in speech \cite{jan2018iwslt}. 
Most NMT systems fall into the category of Sequence-to-Sequence (Seq2Seq) model \cite{sutskever2014sequence}, because both the input and output consist of sequential tokens. 
Therefore, in most neural speech translation, such as that of \cite{bojar2018findings}, the input to the translation system is usually the 1-best hypothesis from the ASR instead of the word lattice output with its corresponding probability scores. 

How to consume word lattice rather than sequential input has been substantially researched in several natural language processing (NLP) tasks, such as language modeling \cite{buckman2018neural}, Chinese Named Entity Recognition (NER) \cite{zhang2018chinese}, and NMT \cite{su2017lattice}.
Additionally, some pioneering works \cite{adams2016learning,sperber2017neural,osamura2018using} demonstrated the potential improvements in speech translation by leveraging the additional information and uncertainty of the packed lattice structure produced by ASR acoustic model.

\begin{figure}[t]
\centering
\includegraphics[width=\columnwidth]{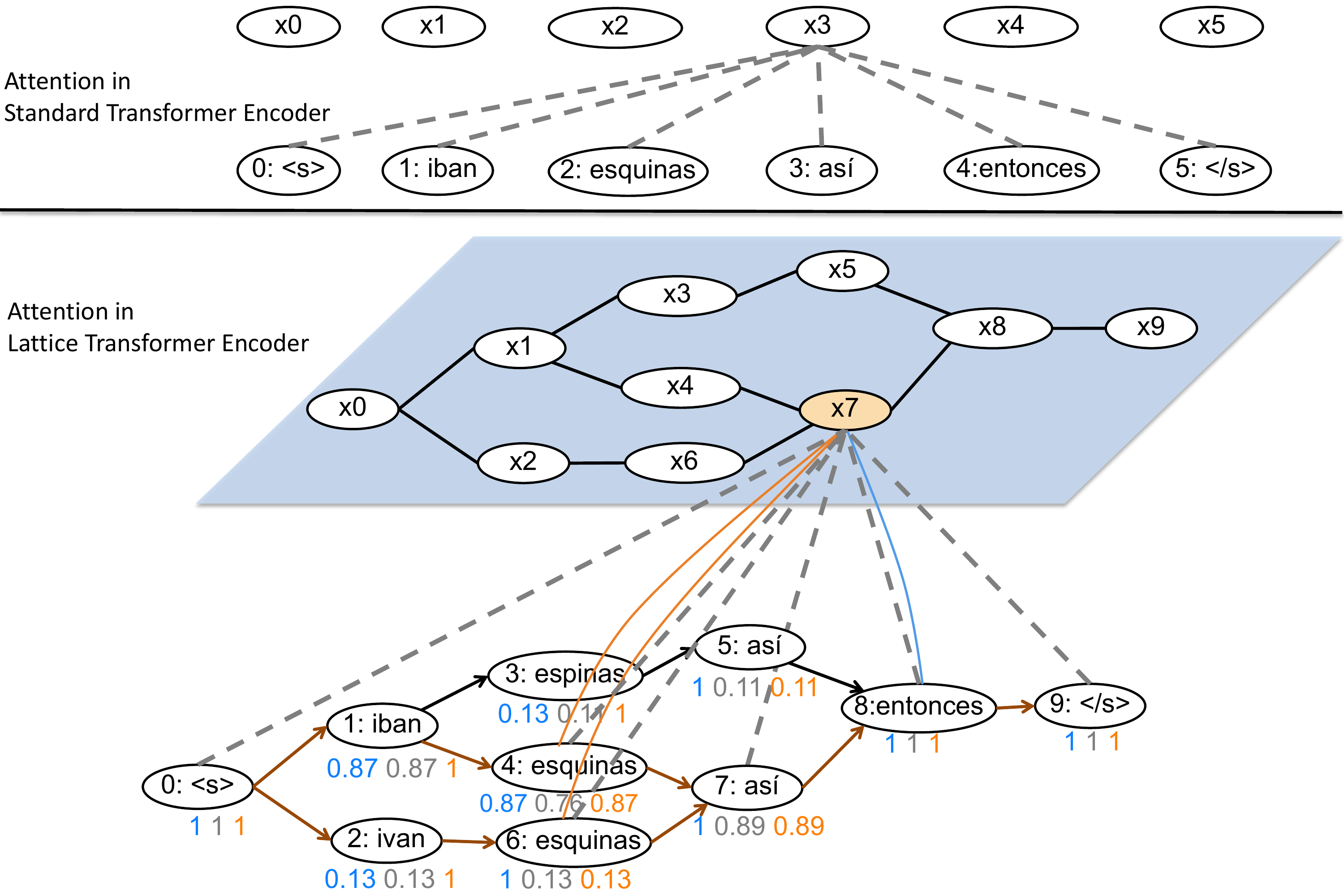}
\caption{Illustration of our proposed attention mechanism (best viewed in color). Our attention depends on the tokens of common paths and forward (blue) / marginal (grey) / backward (orange) probability scores.}
\label{fig:intro}
\end{figure}

Efforts have since continued to push the boundaries of long short-term memory (LSTM) \cite{hochreiter1997long} models. 
More precisely, most previous works are in line with the existing method Tree-LSTMs \cite{tai2015improved}, adapting to task-specific variant Lattice-LSTMs that can successfully handle lattices and robustly establish better performance than the original models. 
However, the inherently sequential nature still remains in Lattice-LSTMs due to the topological representation of the lattice graph, precluding long-path dependencies \cite{khandelwal2018sharp} and parallelization within training examples that are the fundamental constraint of LSTMs. 

In this work, we introduce a generalization of the standard transformer architecture to accept lattice-structured network topologies. 
The standard transformer is a transduction model relying entirely on attention modules to compute latent representations, e.g., the self-attention requires to calculate the intra-attention of every two tokens for each sequence example. 
Latest works such as \cite{yu2018qanet,devlin2018bert,lample2018phrase,su2018unsupervised} empirically find that transformer can outperform LSTMs by a large margin, and the success is mainly attributed to self-attention. 
In our lattice transformer, we propose a lattice relative positional attention mechanism that can incorporate the probability scores of ASR word lattices. 
The major difference with the self-attention in transformer encoder is illustrated in Figure~\ref{fig:intro}. 

We first borrow the idea from the relative positional embedding \cite{shaw2018self} to maximally encode the information of the lattice graph into its corresponding relative positional matrix. 
This design essentially does not allow a token to pay attention to any token that has not appeared in a shared path. 
Secondly, the attention weights depend not only on the query and key representations in the standard attention module, but also on the marginal / forward / backward probability scores \cite{rabiner1989tutorial,post2013improved} derived from the upstream systems (such as ASR). 
Instead of 1-best hypothesis alone (though it is based on forward scores), the additional probability scores have rich information about the distribution of each path \cite{sperber2017neural}. 
It is in principle possible to use them, for example in attention weights reweighing, to increase the uncertainty of the attention for other alternative tokens. 

Our lattice attention is controllable and flexible enough for the utilization of each score. 
The lattice transformer can readily consume the lattice input alone if the scores are unavailable. 
A common application is found in the Chinese NER task, in which a Chinese sentence could possibly have multiple word segmentation possibilities \cite{zhang2018chinese}. 
Furthermore, different BPE operations \cite{sennrich2016neural} or probabilistic subwords \cite{kudo2018subword} can also bring similar uncertainty to subword candidates and form a compact lattice structure.

In summary, this paper makes the following main contributions. 
\textbf{i)} To our best knowledge, we are the first to propose a novel attention mechanism that consumes a word lattice and the probability scores from the ASR system. 
\textbf{ii)} The proposed approach is naturally applied to both the encoder self-attention and encoder-decoder attention. 
\textbf{iii)} Another appealing feature is that the lattice transformer can be reduced to standard lattice-to-sequence model without probability scores, fitting the text translation task. 
\textbf{iv)} Extensive experiments on speech translation datasets demonstrate that our method outperforms the previous transformer and Lattice-LSTMs. 
The experiment on the WMT 2017 Chinese-English translation task shows the reduced model can improve many strong baselines such as the transformer.

\section{Background}

We first briefly describe the standard transformer that our model is built upon, and then elaborate on our proposed approach in the next section.

\subsection{Transformer}

The Transformer follows the typical encoder-decoder architecture using stacked self-attention, point-wise fully connected layers, and the encoder-decoder attention layers. 
Each layer is in principle wrapped by a residual connection \cite{he2016deep} and a postprocessing layer normalization \cite{ba2016layer}. 
Although in principle, it is not necessary to mask for self-attentions in the encoder, in practical implementation it is required to mask the padding positions. 
However, self-attention in the decoder only allows positions up to the current one to be attended to, preventing information flow from the left and preserving the auto-regressive property. 
The illegal connections will be masked out by setting as $-10^9$ before the softmax operation. 

\subsection{Dot-product Attention}

Suppose that for each attention layer in the transformer encoder and decoder, we have two input sequences that can be presented as two matrices $X \in \mathcal{R}^{n \times d}$ and $Y \in \mathcal{R}^{m \times d}$, where $n,m$ are the lengths of source and target sentences respectively, and $d$ is the hidden size (usually equal to embedding size), the output is $h$ new sequences $Z_i \in \mathcal{R}^{n \times d/h}$ or $\in \mathcal{R}^{m \times d/h}$, where $h$ is the number of heads in attention. 
In general, the result of multi-head attention is calculated according to the following procedure.
\begin{align}
Q &= X W^Q \text{ or } Y W^Q \text{ or } Y W^Q \label{eq:self_q} \\
K &= X W^K \text{ or } Y W^K \text{ or } X W^K \label{eq:self_k} \\
V &= X W^V \text{ or } Y W^V \text{ or } X W^V \label{eq:self_v} \\
Z_i &= \textbf{Softmax}\left(\frac{Q K^\top}{\sqrt{d/h}} + \mathbb{I}_{d}M \right) V \label{eq:self_dot} \\
Z &= \textbf{Concat}(Z_1,...,Z_h) W^O \label{eq:self_att}
\end{align}
where the matrices $W^Q, W^K, W^V \in \mathcal{R}^{d \times d/h}$ and $W^O \in \mathcal{R}^{d \times d}$ represent the learnable projection parameters, and the masking matrix $M \in \mathcal{R}^{m \times m}$ is an upper triangular matrix with zero on the diagonal and non-zero ($-10^9$) everywhere else.

Note that \textbf{i)} the three columns in the right-side of Eq~(\ref{eq:self_q},\ref{eq:self_k},\ref{eq:self_v}) are used to compute the encoder self-attention, the decoder self-attention, and the encoder-decoder attention respectively, \textbf{ii)} $\mathbb{I}_d$ is the indicator function that returns 1 if it computes decoder self-attention and 0 otherwise, \textbf{iii)} the projection parameters are unique per layer and head,  \textbf{iv)} the \textbf{Softmax} in Eq~(\ref{eq:self_dot}) means a row-wise matrix operation, computing the attention weights by scaled dot product and resulting in a simplex $\Delta^n$ for each row.

\section{Lattice Transformer}

As motivated in the introduction, our goal is to enhance the standard transformer architecture, which is limited to sequential inputs, to consume lattice inputs with additional information from the up-stream ASR systems.

\subsection{Lattice Representation}

\begin{figure}[t]
\centering
\includegraphics[width=\columnwidth]{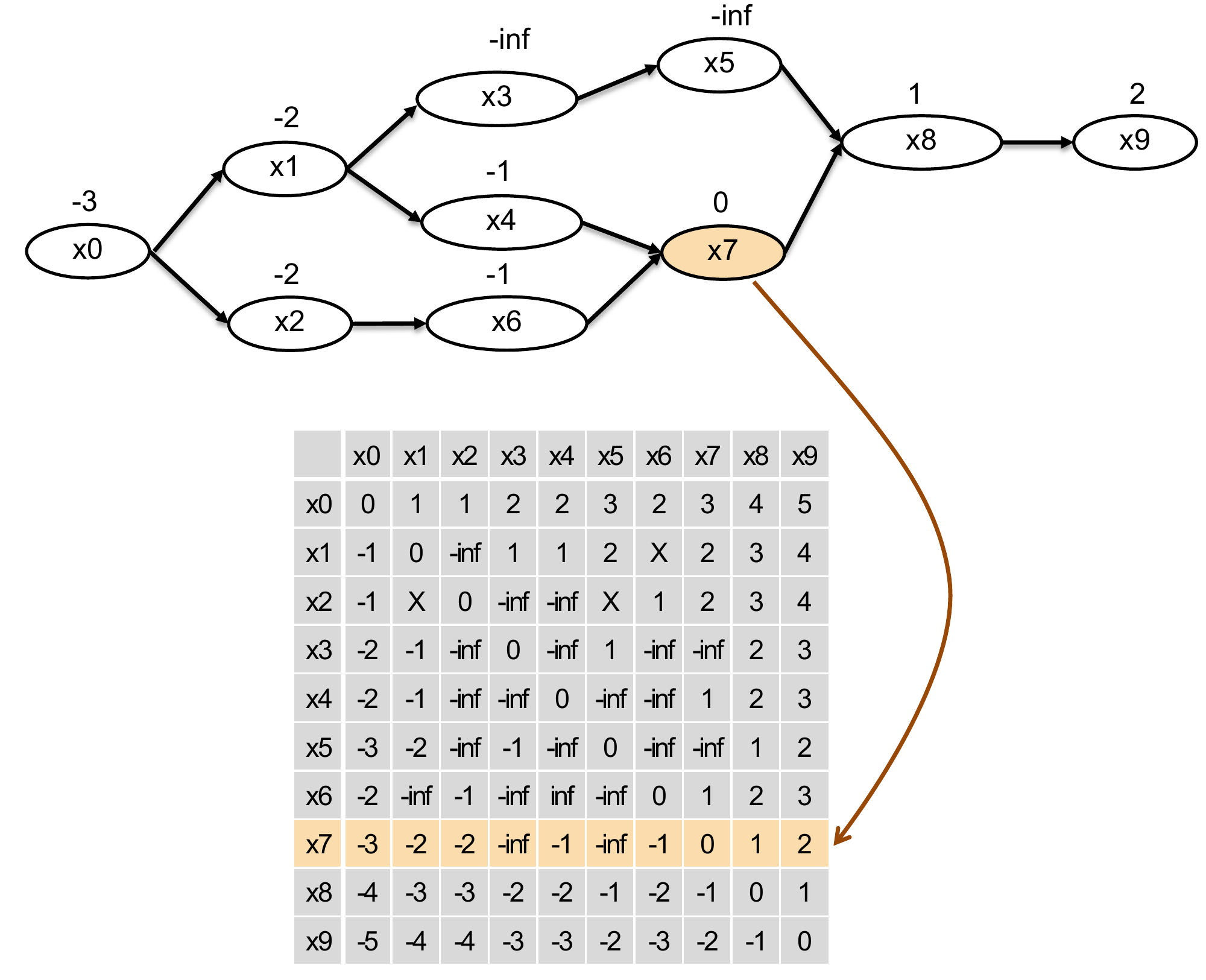}
\caption{An example of the lattice relative position matrix, where ``-inf" in the matrix is a special number denoting that no relative position exists between the corresponding two tokens.}
\label{fig:lat_mat}
\end{figure}

Without loss of generality, we assume a word lattice from ASR system to be a directed, connected and acyclic graph following a topological ordering such that a child node comes after its parent nodes. 
We add two special tokens to each path of the lattice, which represent the start of sentence and the end of sentence (e.g., Figure~\ref{fig:intro}), so that the graph has a single source node and a single end node, where each node is assigned a token. 

Given the definition and property described above, we propose to use a relative positional \textbf{lattice matrix} $L \in \mathcal{N}^{n \times n}$ to encode the graph information, where $n$ is number of nodes in the graph.  
For any two nodes $i,j$ in the lattice graph, the matrix entry $L_{ij}$ is the minimum relative distance between them. 
In other words, if the nodes $i,j$ share at least one path, then we have
\begin{equation} 
L_{ij} = \min_{p\in\text{common paths for $i,j$}}L_{i0}^p - L_{j0}^p ,\label{eq:lat_mat}
\end{equation}
where $L_{\cdot0}^p$ is the distance to the source node in path $p$. 
If no common path exists for two nodes, we denote the relative distance as $-\infty$ ($-10^9$ in practice) for subsequent masking in the lattice attention. 
The reason for choosing the ``$\min$" in Eq~(\ref{eq:lat_mat}) is that in our dataset we find about 70\% of $L_{ij}$s computed by ``$\min$" and ``$\max$" are identical, and about 20\% entries just differ by 1. 
Empirically, our experiments also show no significant difference in the performance of either one.

An illustration to compute the lattice matrix for the example in the introduction is shown in Figure~\ref{fig:lat_mat}.  
Since we can deterministically reconstruct the lattice graph from those matrix elements that are equal to 1, it indicates the relation information between the parent and child nodes. 

\subsection{Controllable Lattice Attention}

\begin{figure*}[t]
\centering
\includegraphics[width=0.475\textwidth]{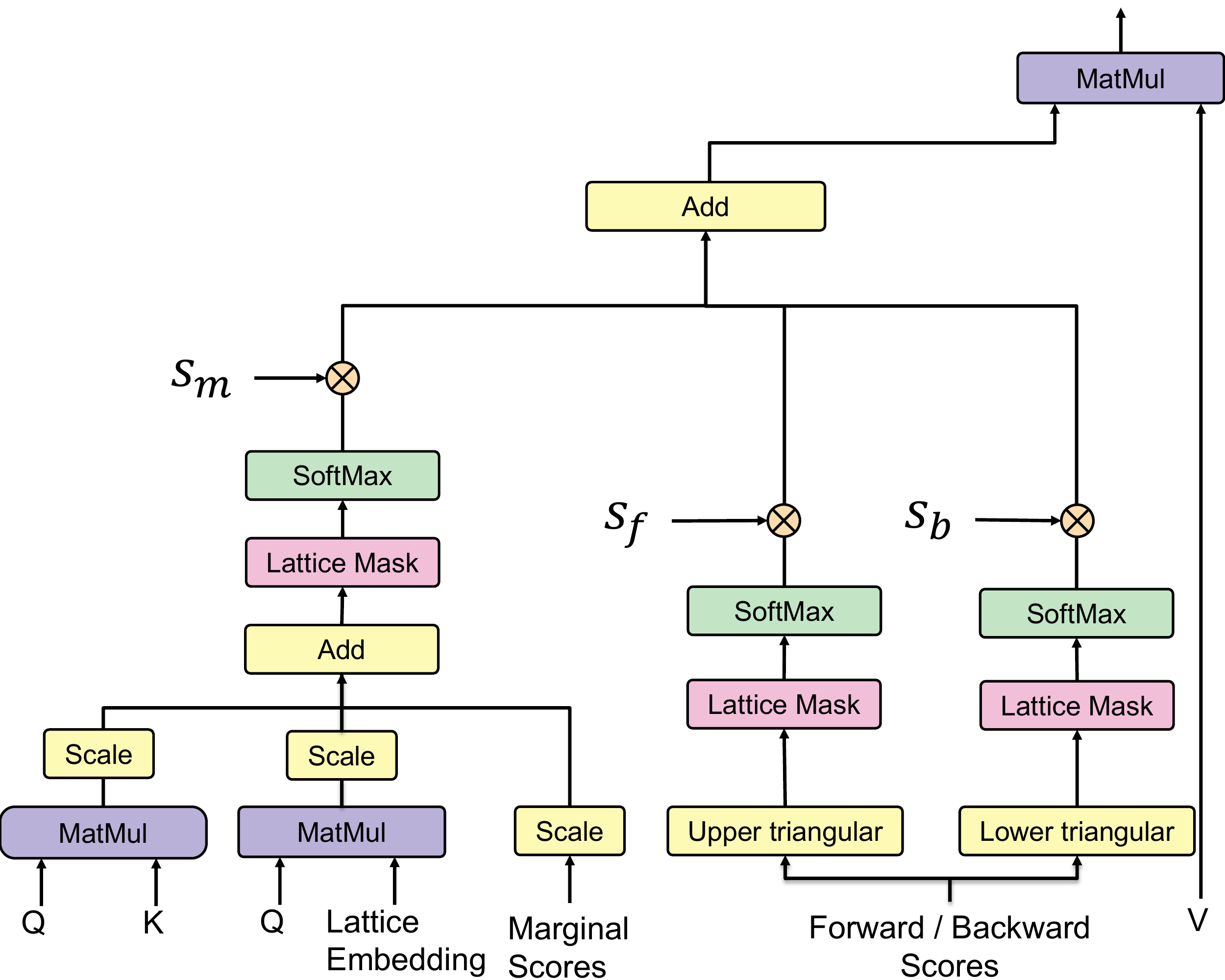}
\includegraphics[width=0.475\textwidth]{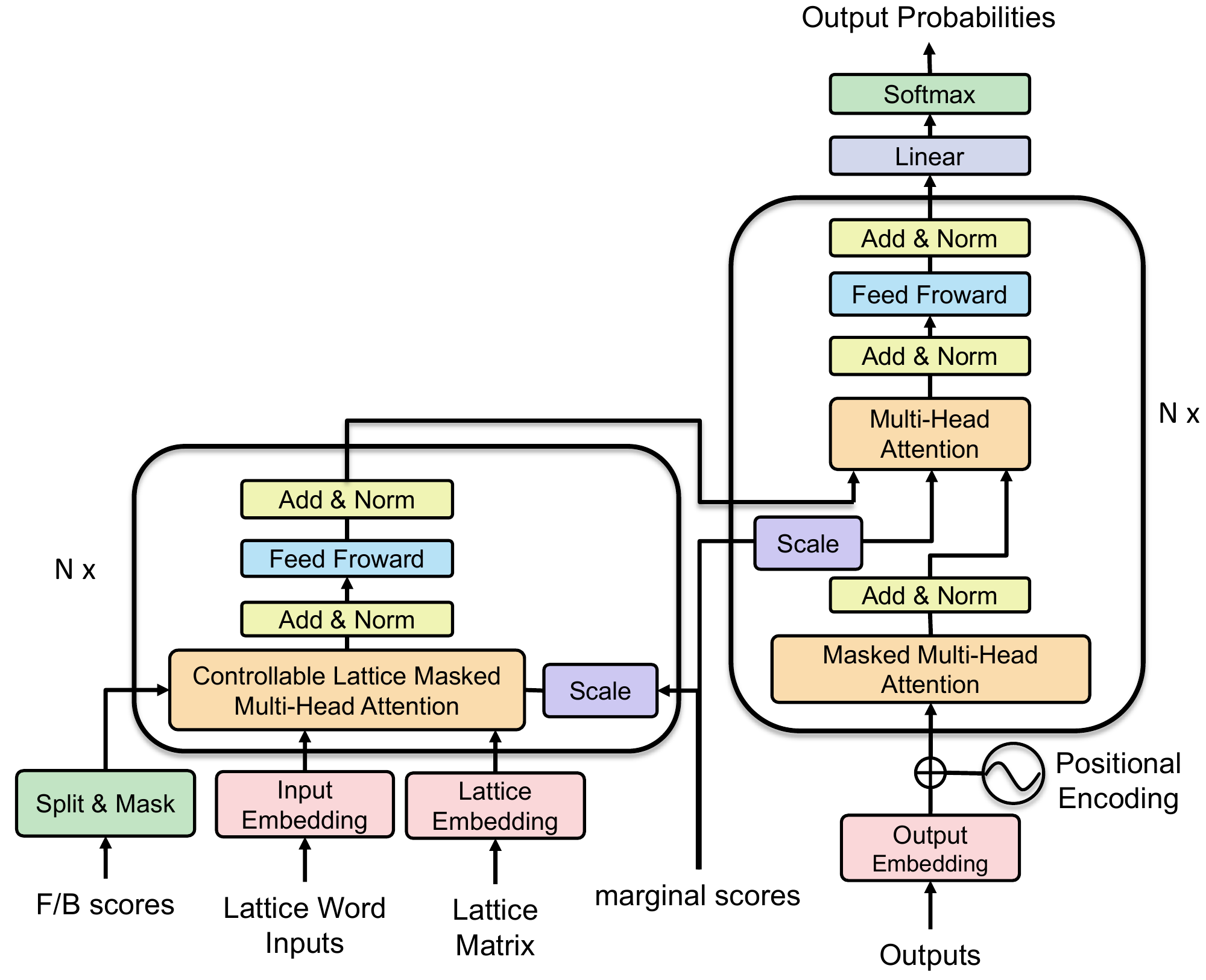}
\caption{Left panel: the controllable lattice attention, where $s_m,s_f,s_b$ are learnable scalars and $s_m+s_f+s_b=1$. Right panel: the overall model architecture of lattice transformer.}
\label{fig:lat_att}
\end{figure*}

Besides the lattice graph representation, the posterior probability scores can be simultaneously produced from the acoustic model and language model in most ASR systems. 
We deliberately design a controllable lattice attention mechanism to incorporate such information to make the attention encode more uncertainties. 

In general, we denote the posterior probability of a node $i$ as the forward score $f_i$, where the summation of the forward scores for its child nodes is 1. 
Following the recursion rule in \cite{rabiner1989tutorial}, we can further derive another two useful probabilities, the marginal score $m_i = f_i\sum_{j\in\text{Pa}(i)} m_j$ and the backward score $b_i=m_i/\sum_{k\in\text{Ch}(i)}m_k$, where $\text{Pa}(i)$ or $\text{Ch}(i)$ denotes node $i$'s predecessor or successor set, respectively. 
Intuitively, the marginal score measures the global importance of the current token compared with its substitutes given all predecessors; the backward score is analogous to the forward score, which is only locally associated with the importance of different parents to their children, where the summation of its parent nodes' scores is 1.
Therefore, our controllable attention aims to employ marginal scores and forward / backward scores. 

\subsubsection{Lattice Embedding}

We first construct the latent representations of the relative positional lattice matrix $L$. 
The matrix $L$ can be straightforwardly decomposed into two matrices: one is the mask $L^M$ with only 0 and $-\infty$ values, and the other is the matrix with regular values i.e., $L^R = L - L^M$. 
Given a 2D embedding matrix $W^L$, the embedded vector of $L^R_{ij}$ can be written as $W^L[L^R_{ij},:]$ with the NumPy style indexing. 
In order to prevent the the lattice embedding from dynamically changing, we have to clip every entry of $L^R$ with a positive integer $c$\footnote{$\text{clip}(l, c)=\max(-c, \min(l,c))$}, such that $W^L \in \mathcal{R}^{(2c+1)\times d/h}$ has a fixed dimensionality and becomes learnable parameters. 

\subsubsection{Attention with Probability Scores}

Our proposed controllable lattice attention is depicted in the left panel of Figure~\ref{fig:lat_att}. It shows the computational graph with detailed network modules. 
More concretely, we first denote the lattice embedding for $L^R$ as a 3D array $E \in \mathcal{R}^{n \times n \times d/h}$. 
Then, the attention weights adapted from traditional transformer are integrated with marginal scores that capture the distribution of each path in the lattice. 
The logits in Eq~(\ref{eq:self_dot}) will become the addition of three individual terms (if we temporarily omit the mask matrix), 
\begin{align}
&\frac{QK^\top + \textbf{einsum}(\texttt{'ik,ijk->ij'}, Q, E)}{\sqrt{d/h}} \nonumber \\
&+ w_m\mathbf{m} ~. \label{eq:lat_dot}
\end{align}
The original $QK^\top$ will remain since the word embeddings have the majority of significant semantic information. 
The difficult part in Eq~(\ref{eq:lat_dot}) is the new dot product term involving the lattice embedding by \textbf{einsum}\footnote{This op is available in NumPy, TensorFlow, or PyTorch. In our example, $Q$ and $E$ are 2D and 3D arrays, and the result of this op is a 2D array, with the element in $i$th row, $j$th column is $\sum_{k}Q_{ik}E_{ijk}$.} operation, where \textbf{einsum} is a multi-dimensional linear algebraic array operation in Einstein summation convention. 
In our case, it tries to sum out the dimension of the hidden size, resulting in a new 2D array $\in \mathcal{R}^{n \times n}$, which is further be scaled by $\frac{1}{\sqrt{d/h}}$ as well. 
In addition, we aggregate the scaled marginal score vector $\mathbf{m} \in \mathcal{R}^n$ together to obtain the logits. 
With the new parameterization, each term has an intuitive meaning: term i) represents semantic information, term ii) governs the lattice-dependent positional relation, term iii) encodes the global uncertainty of the ASR output.

The attention logits associated with the forward or backward scores are much different from marginal scores, since they govern the local information between the parent and child nodes. 
They are represented as a matrix rather than a vector, where the matrix has only non-zero values if nodes $i,j$ have a parent-child relation in the lattice graph.  
First, an upper or lower triangular mask matrix is used to enforce every token's attention to the forward scores of its successors or the backward scores of its predecessors. 
It seems counter-intuitive but the reason is that the summation of the forward scores for each token's child nodes is 1. 
So is the backward scores of each token's parent nodes. 
Secondly, before applying the softmax operation, the lattice mask matrix $L^M$ is added to each logits to prevent attention from crossing paths.  
Eventually, the final attention vector used to multiply the value representation $V$ is a weighed averaging of the three proposed attention vectors $A_{\cdot}$ with different probability scores $s_{\cdot}$,
\begin{align}
A_{final} =& s_m A_m + s_f A_f + s_b A_b, \label{eq:final_att}\\
\text{s.t.~~~~} & s_m+s_f+s_b=1~. \notag
\end{align}
In summary, the overall architecture of lattice transformer is illustrated in the right of Figure~\ref{fig:lat_att}.

\subsubsection{Discussion}

A critical point for the lattice transformer is whether the model can generalize to other common lattice-based inputs. 
More specifically, how does the model apply to the lattice input without probability scores? And to what extent can we train the lattice model on a regular sequential input?
If probability scores are unavailable, we can use the lattice graph representations alone by setting the scalar $w_m=0$ in Eq~(\ref{eq:lat_dot}) and $s_f=s_b=0$, $s_m=1$ in Eq~(\ref{eq:final_att}) as non-trainable constants. 
We validate this viewpoint on the Chinese-English translation task, where the Chinese input is a pure lattice structure derived from different tokenizations. 
As to sequential inputs, it is just a special case of the lattice graph with only one path. 

An interesting point to mention is that our encoder-decoder attention also takes the key and value representations from the lattice input and aggregates the marginal scores, though the sequential target forbids us to use lattice self-attention in the decoder. 
However, we can still visualize how the sequential target attends to the lattice input.

A practical point for the lattice transformer is whether the training or inference time for such a seemingly complicated architecture is acceptable. 
In our implementation, we first preprocess the lattice input to obtain the position matrix for the whole dataset, thus the one-time preprocessing will bring almost no over-head to our training and inference. 
In addition, the extra \textbf{enisum} operation in controllable lattice attention is the most time-consuming computation, but remaining the same computational complexity as $QK^\top$. 
Empirically, in the ASR experiments, we found that the training and inference of the most complicated lattice transformer (last row in the ablation study) take about 100\% and 40\% more time than standard transformer; in the text translation task, our algorithm takes about 30\% and 20\% more time during training and inference.

\section{Experiments}\

We mainly validate our model in two scenarios, speech translation with word lattices and posterior scores, and Chinese to English text translation with different BPE operations on the source side. 

\subsection{Speech Translation}

For the speech translation experiment, we use the Fisher and Callhome Spanish-English Speech Translation Corpus from LDC \cite{post2013improved}, which is produced from telephone conversations. 
Our baseline models are the vanilla Transformer with relative positional embeddings \cite{vaswani2017attention,shaw2018self}, and Lattice-LSTMs \cite{sperber2017neural}. 

\subsubsection{Datasets}

The Fisher corpus includes the contents between strangers, while the Callhome corpus is primarily between friends and family members. 
The numbers of sentence pairs of the two datasets are respectively 138,819 and 15,080. 
The source side Spanish corpus consists of four data types: \textbf{reference} (human transcripts), \textbf{oracle} of ASR lattices (the optimal path with the lowest word error rate (WER)), \textbf{ASR 1-best} hypothesis, and \textbf{ASR lattice}. 
For the data processing, we make case-insensitive tokenization with the standard moses\footnote{https://github.com/moses-smt/mosesdecoder} tokenizer for both the source and target transcripts, and remove the punctuation in source sides. 
The sentences of the other three types have been already been lowercased and punctuation-removed. 
To keep consistent with the lattices, we add a token ``$<$s$>$" at the beginning for all cases. 

\begin{table}[h]
\setlength{\tabcolsep}{2pt}
\centering
\resizebox{0.8\columnwidth}{!}{
\begin{tabular}{c|c}
Setting & Description \\
\hline
R & baseline, trained with human transcripts only\\
R+1 & fine-tuned on 1-best hypothesis\\
R+L & fine-tuned on lattices without probability scores\\
R+L+S & fine-tuned on lattices with probability scores\\
\hline
\end{tabular}
}
\caption{4 systems for comparison}
\label{tab:setting}
\end{table}

\begin{table*}[ht]
\centering
\resizebox{\textwidth}{!}{
\begin{tabular}{|l|c|c|c|c|c|c|c|c|c|c|c|c|c|}
\hline
\multirow{2}{*}{Architecture} & 
\multirow{2}{*}{Inference Inputs} &
\multicolumn{4}{|c|}{Fisher dev} &
\multicolumn{4}{|c|}{Fisher dev2}&
\multicolumn{4}{|c|}{Fisher test}\\
\cline{3-14}
& & R & R+1 & R+L & R+L+S & R & R+1 & R+L & R+L+S & R & R+1 & R+L & R+L+S\\
\hline
\multirow{3}{*}{Lattice LSTM} 
& reference & - &- &- &- &53.9 & 53.8 & 53.7 & 54 &52.2 & 51.8 & 52.2 & 52.7\\
\cline{2-14}
& oracle & - &- &- &- &  44.9&	45.6	&45.2&	45.2&	44.4&	44.6	&44.6&	44.8\\
\cline{2-14}
& ASR 1-best & - &- &- &- &  35.8&	37.1&	36.2&	36.2&	35.9&	36.6&	36.2&	36.4\\
\cline{2-14}
& ASR Lattice & - & -&	-&-&	25.9&	25.8&	36.9&	38.5&	26.2&	25.8&	36.1&	38\\
\hline
\multirow{3}{*}{Lattice Transformer} 
& reference &57.1&	55.0&55.5	&	55.5&	58.0&	56.1&56.4	&	56.6&	56.0&	53.7&54.1	&	54.2\\
\cline{2-14}
& oracle & 46.3&	46.2&46.8	&	46.7&	47.1&	47.0&47.5	&	47.9&	46.8&	46.4&46.9	&	46.9\\
\cline{2-14}
& ASR 1-best & 36.5&	37.4&37.6	&	37.4&	37.4&	\textcolor{blue}{38.4} & 38.3	&	38.6&	37.7&	\textcolor{blue}{38.5} &38.2	&	38.4\\
\cline{2-14}
& ASR Lattice& 32.9&	33.8&37.7	&	\textbf{38.3}&	33.4&	34.0& \textcolor{orange}{38.6} &	\textbf{39.4}&	33.5&	33.7& \textcolor{orange}{37.9}	&	\textbf{39.2}\\
\hline
\end{tabular}
}
\caption{Cross-Evaluation of BLEU on Fisher. Note that for the lattice transformer architecture with R or R+1 setting, the resulted model is equivalent to a standard transformer with relative positional embeddings. The evaluation of \textbf{oracle} inputs is similar to ASR 1-best, but it can indicate an upper bound of the performance. The evaluation results of Lattice LSTM on Fisher dev are not reported in \cite{sperber2017neural}.}
\label{table_fisher}
\end{table*}

\begin{table}[ht]
\centering
\resizebox{\columnwidth}{!}{
\begin{tabular}{|l|c|c|c|c|c|}
\hline
\multirow{2}{*}{Architecture} & 
\multirow{2}{*}{Inference Inputs} &
\multicolumn{4}{|c|}{Callhome devtest} \\
\cline{3-6}
& & R & R+1 & R+L & R+L+S \\
\hline
\multirow{4}{*}{Lattice Transformer} 
& reference & 28.3&	29.6&	30.0&	30.4  \\
\cline{2-6}
& oracle & 17.7 & 19.7 &	19.5&	19.6\\
\cline{2-6}
& ASR 1-best & 13.4&	15.2&	14.8&	15.1\\
\cline{2-6}
& ASR Lattice&  13.4&	13.4&	15.6&	\textbf{15.7}\\
\hline
\multirow{2}{*}{} & 
\multirow{2}{*}{} &
\multicolumn{4}{|c|}{Callhome evltest} \\
\cline{3-6}
& & R & R+1 & R+L & R+L+S \\
\hline
\multirow{4}{*}{Lattice LSTM} 
& reference & 24.7&	24.3&	24.8&	24.4\\
\cline{2-6}
& oracle & 15.8&	16.8&	16.3&	15.9\\
\cline{2-6}
& ASR 1-best & 11.8&	13.3&	12.4&	12.0\\
\cline{2-6}
& ASR Lattice & 9.3&	7.1&	13.7&	14.1\\
\hline
\multirow{4}{*}{Lattice Transformer} 
& reference & 27.1&	28.6&	28.9&	29.1\\
\cline{2-6}
& oracle & 16.5 & 18.1&	17.7&	18.0\\
\cline{2-6}
& ASR 1-best & 12.7&	\textcolor{blue}{14.5}&	13.6&	14.1\\
\cline{2-6}
& ASR Lattice& 12.7&	13.0&\textcolor{orange}{14.2}&\textbf{14.9}\\
\hline
\end{tabular}
}
\caption{Cross-Evaluation of BLEU on Callhome.}
\label{table_callhome}
\end{table}

\subsubsection{Training and Cross-Evaluation} 

4 systems in Table~\ref{tab:setting} are trained for both Lattice-LSTMs and Lattice Transformer. 
For fair and comprehensive comparison, we also evaluate all algorithms on the inputs of four types. 
We initially train the baseline of our lattice transformer with the human transcripts on Fisher/Train data alone, which is equivalent to the modified transformer \cite{shaw2018self}. 
Then we fine-tune the pre-trained model with 1-best hypothesis or word lattices (and probability scores) for either Fisher or Callhome dataset. 

The source and target vocabularies are built respectively from the transcripts of Fisher/Train and Callhome/Train corpus, with vocabulary sizes 32000 and 20391. 
The hyper-parameters of our model are the same as Transformer-base with 512 hidden size, 6 attention layers, 8 attention heads and beam size 4. 
We use the same optimization strategy as \cite{vaswani2017attention} for pre-training with 4 GPU cards, and apply SGD with constant learning rate 0.15 for fine-tuning. 
We select the best performed model based on Fisher/Dev or Callhome/Dev, and test on Fisher/Dev2, Fisher/Test or Callhome/Test. 

To better analyze the performance of our approach, we use an intensive cross-evaluation method, i.e., we feed 4 possible inputs to test different models. 
The cross-evaluation results are put into several $4 \times 4$ blocks in Table~\ref{table_fisher} and \ref{table_callhome}. 
As the aforementioned discussion, if the input is not ASR lattice, the evaluation on the model R+L+S needs to set $w_m=s_f=s_b=0,s_m=1$. 
If the input is an ASR lattice but fed into the other three models, the probability scores are in fact discarded.   

\begin{table*}[t]
\setlength{\tabcolsep}{2pt}
\centering
\resizebox{\textwidth}{!}{
\begin{tabular}{l|l|l|l|l}
\hline
src transcript & qu\'{e} tal , eh , yo soy guillermo , ?` c\'{o}mo est\'{a}s ? & porque como esto tiene que ir avanzando ?` no ? & pues , ?` y llevas muchos a\~{n}os aqu\`{i} en atlanta ? & quererlo y tener fe . \\
\hline 
tgt reference & how are you , eh i 'm guillermo , how are you ? & because like this has to be moving forward , no ? &  well . and you 've been many years here in atlanta ? & to love him and have faith .\\
\hline
ASR 1-best & quedar eh yo soy guillermo c\'{o}mo est\'{a}s & porque como esto tiene que ir avanzando no & país lleva muchos años aquí en atlanta & quieren lo y tener fe \\
\hline
mt from R+1 & stay . eh , i 'm guillermo . how are you ? & why do you have to move forward or not ? & country has been many years here in atlanta & they want to have faith \\
\hline
ASR lattice & \tabincell{l}{quedar que qu\'{e} eh yo soy dar eh yo tal eh yo soy \\ guillermo cómo comprar con como est\'{a} est\'{a}s} & porque como esto tiene que ir avanzando no & pa\'{i}s well lleva lleva muchos a\~{n}os aqu\'{i} en atlanta & \tabincell{l}{quieren quererlo lo y tener \\ tenerse fe y tener tenerse fe} \\
\hline
mt from R+L+S & how are you ? i 'm guillermo . how are you ? & because since this has to move forward , right ? & well , you 've been here many years in atlanta & loving him and having faith\\
\hline
\end{tabular}
}
\caption{Translation examples on test sets. Note that the presented ASR lattice does not include lattice information.}
\label{tab:speech_trans}
\end{table*}

\subsubsection{Results on Fisher and Callhome}

We mainly compare our architecture with the previous Lattice-LSTMs \cite{sperber2017neural} and the transformer \cite{shaw2018self} in Table~\ref{table_fisher}.
Since the transformer itself is a powerful architecture for sequence modeling, the BLEU scores of the baseline (R) have significant improvement on test sets. 
In addition, fine-tuning without scores hasn't outperformed the 1-best hypothesis fine-tuning, but has about 0.5 BLEU improvement on oracle and transcript inputs. 
We suspect this may be due to the high ASR WER and if the ASR system has a lower WER, the lattice without score fine-tuning may get a better translation. 
We will leave this as a future research direction on other datasets from better ASR systems. 
For now, we just validate this argument in the BPE lattice experiments, and detailed discussion sees next section. 
As to fine-tuning with both lattices and probability scores, it increases the BLEU with a relatively large margin of 0.9/1.0/0.7 on Fisher Dev/Dev2/Test sets. 
Besides, for ASR 1-best inputs, it is still comparable with the R+1 systems, while for oracle and transcript inputs, there are about 0.5-0.9 BLEU score improvements.

The results of Callhome dataset are all fine-tuned from the pre-trained model based on Fisher/Train corpus, since the data size of Callhome is too small to train a large deep learning model. 
This is the reason why we adopt the strategy for domain adaption. 
We use the same method for model selection and test. 
The detailed results in Table~\ref{table_callhome} show the consistent performance improvement.

\begin{figure}[t]
\centering
\includegraphics[width=0.4\textwidth]{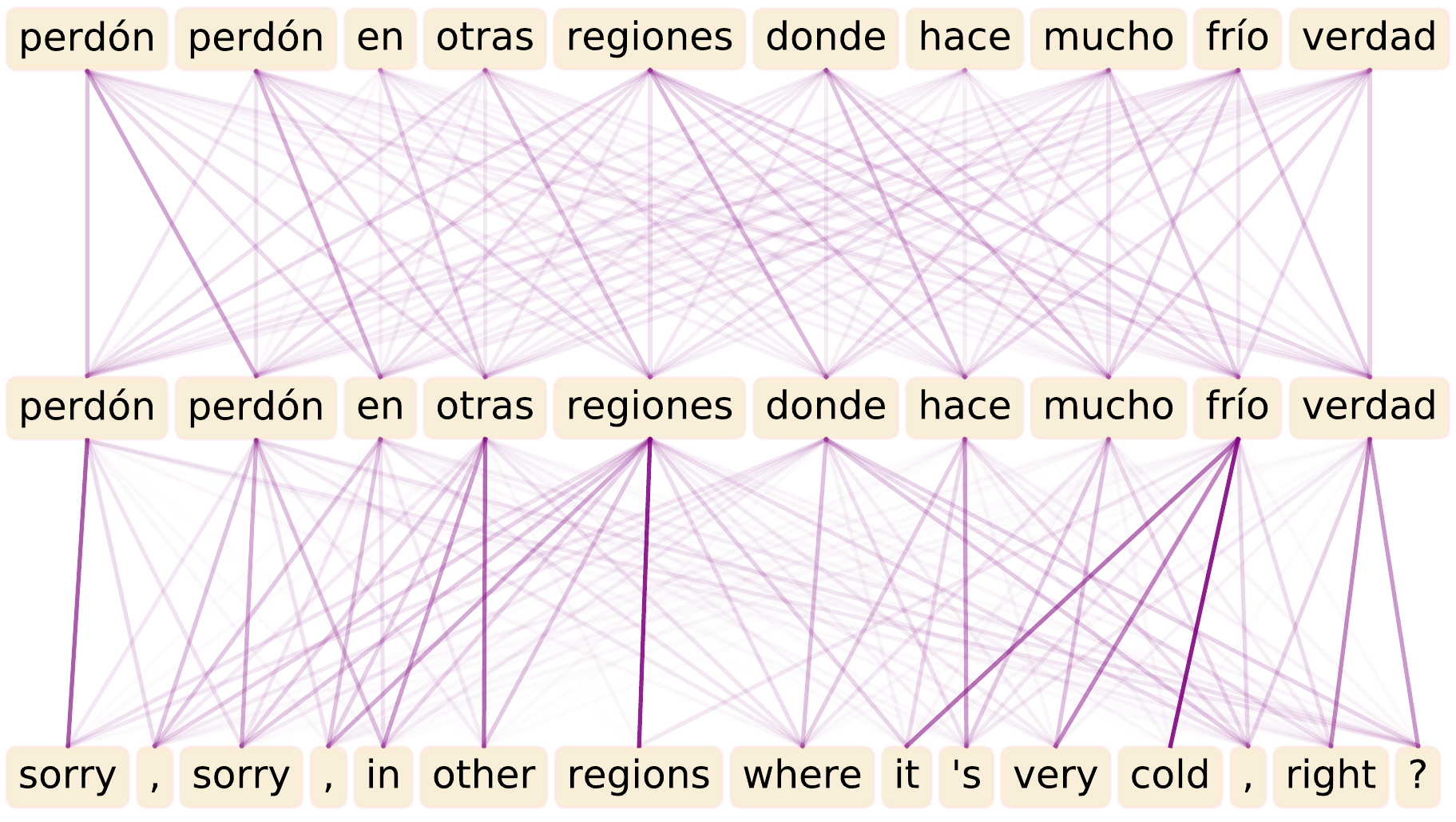}
\includegraphics[width=0.4\textwidth]{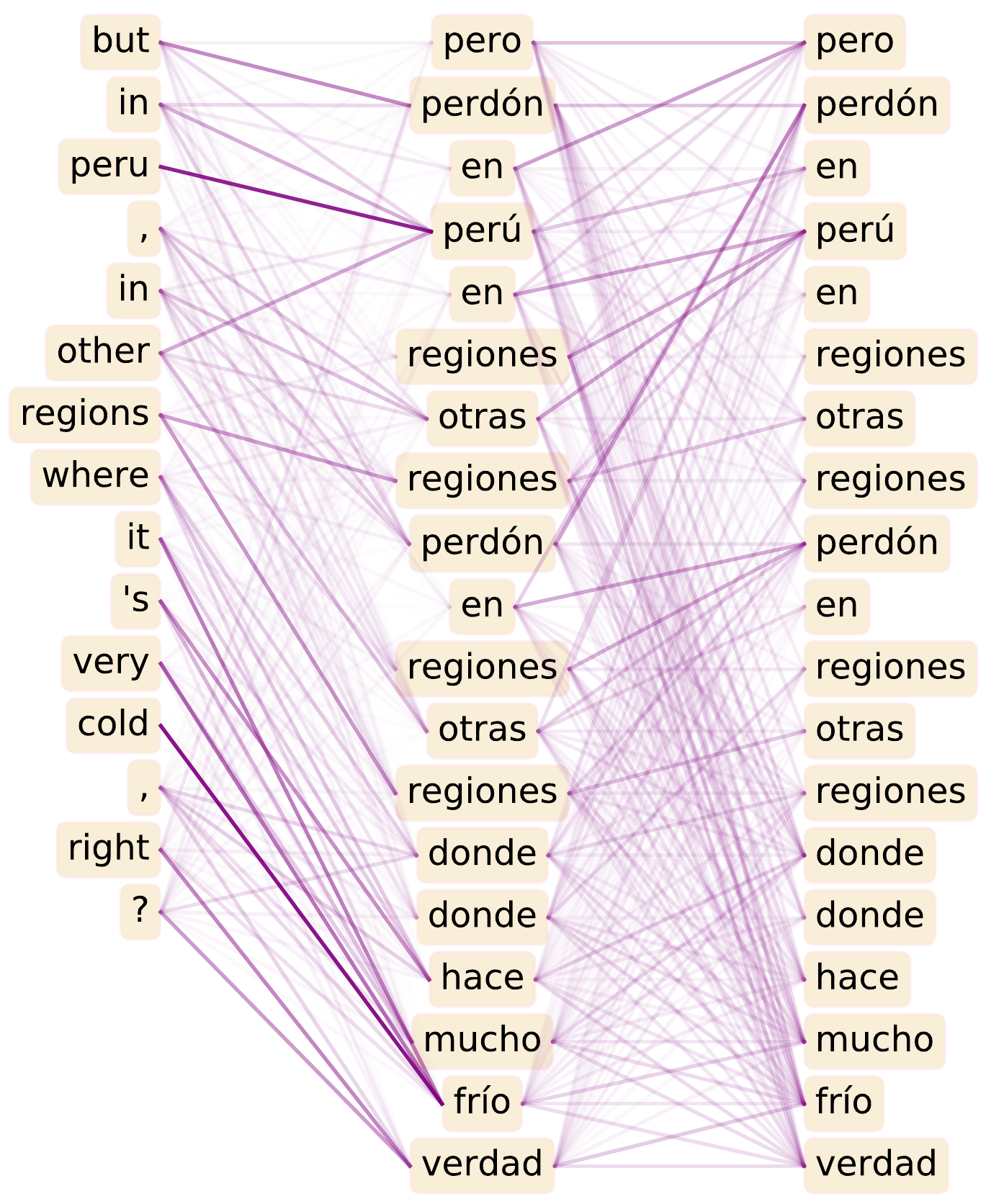}
\caption{Visualization of Lattice Transformer encoder self-attention and encoder-decoder attention for inference. Top panel: ASR 1-best. Bottom panel: ASR lattice. Target reference: ``But in Peru, I've heard there are parts where it really gets cold."}
\label{fig:fisher204_visual}
\end{figure}

\begin{figure}[t]
\centering
\includegraphics[width=0.137\textwidth]{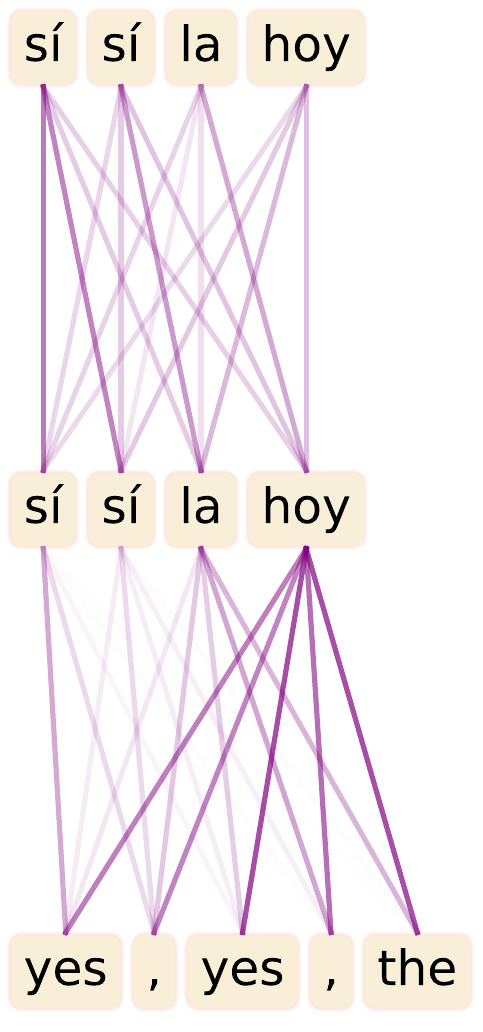}
\includegraphics[width=0.21\textwidth]{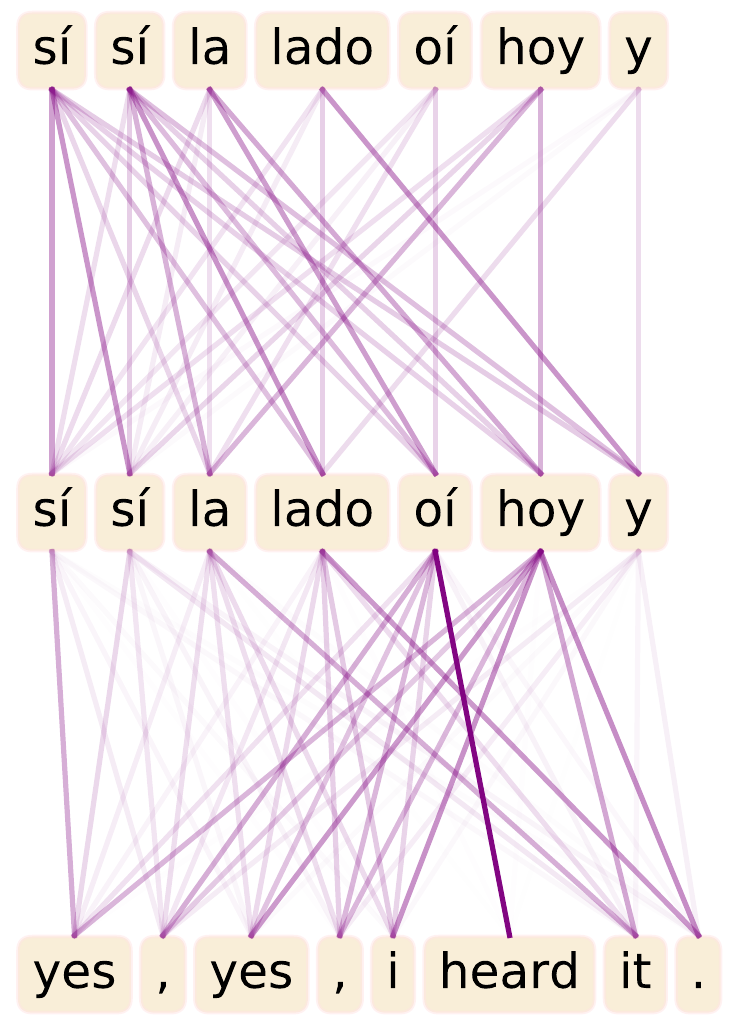}
\caption{Left panel: ASR 1-best. Right panel: ASR lattice. Target reference: ``Yes, yes, I heard it."}
\label{fig:callhome94_visual}
\end{figure}

\subsubsection{Inference Analysis}

On the test datasets of Fisher and Callhome, we make an inference for predicting the translations, and some examples are shown in Table~\ref{tab:speech_trans}. 
We also visualize the alignment for both encoder self-attention and encoder-decoder attention for the input and predicted translation. 
Two examples are illustrated in Figure~\ref{fig:fisher204_visual} and~\ref{fig:callhome94_visual}. 
As expected, the tokens from different paths will not attend to each other, e.g., ``pero" and ``perd\'{o}n" in Figure~\ref{fig:fisher204_visual} or ``hoy" and ``y" in Figure~\ref{fig:callhome94_visual}. 
In Figure~\ref{fig:fisher204_visual}, we observe that the 1-best hypothesis can even result in erroneous translation ``sorry, sorry", which is supposed to be ``but in peru". 
In Figure~\ref{fig:callhome94_visual}, the translation from 1-best hypothesis obviously misses the important information ``i heard it".  
We primarily attribute such errors to the insufficient information within 1-best hypothesis, but if the lattice transformer is appropriately trained, the translations from lattice inputs can possibly correct them. 
Due to limited space, more visualization examples can be found in supplementary material.

\begin{table}[t]
\setlength{\tabcolsep}{2pt}
\centering
\resizebox{\columnwidth}{!}{
\begin{tabular}{lccc}
\hline
Model & 
Fisher dev2 &
Fisher test &
Callhome evltest \\
\hline
LSTM (1-best input) & 37.1&	36.6&	13.3\\
Lattice LSTM (lattice input) &	36.9&	36.1&	13.7\\
+lattice prob scores&	38.5&	38&	14.1\\
\hline
Transformer (1-best input) &\textcolor{blue}{38.4}&	\textcolor{blue}{38.5}&	\textcolor{blue}{14.5}\\
Lattice Transformer (lattice input) &\textcolor{orange}{38.6}&	\textcolor{orange}{37.9}&	\textcolor{orange}{14.2}\\
+ marginal scores in decoder&	39.0&	38.7&	14.4\\
+ marginal scores in encoder&	38.8&	38.2&	14.7\\
+ marginal scores in encoder and decoder&	39.5&	39.0&	14.8\\
\tabincell{l}{+ marginal scores in encoder and decoder,\\ and forward / backward scores only \\ in encoder self-attention layer 0 and layer 1}&		\textbf{39.6}&	39.1&		\textbf{14.9}\\
\tabincell{l}{+ marginal scores in encoder and decoder, \\and forward / backward scores \\ in all encoder self-attention layers}&	39.4 &		\textbf{39.2}&	14.7\\
\hline
\end{tabular}
}
\caption{Ablation Experiment BLEU Results. The rows of the Lattice LSTM and the Lattice Transformer represent the 1-best hypothesis fine-tuning, and the BLEUs are evaluated on 1-best inputs and on lattice inputs for the others. The colored BLEU values come from Table~\ref{table_fisher} and~\ref{table_callhome}.}
\label{table_ablation}
\end{table}

\subsubsection{Model Ablation Study}

We conduct an ablation study to examine the effectiveness of every module in the lattice transformer. 
We gradually add one module from a standard transformer model to our most complicated lattice transformer. 
From the results in Table~\ref{table_ablation}, we can see that the application of marginal scores in encoder or decoder has the most influential impact on the lattice fine-tuning. 
Furthermore, the superimposed application of marginal scores in both encoder and decoder can gain an additional promotion, compared to individual applications. 
However, the use of forward and backward scores has no significantly extra rewards in this situation. 
Perhaps due to overfitting, the most complicated lattice transformer on the Callhome of smaller data size cannot achieve better BLEUs than simpler models.

\begin{figure}[h!]
\centering
\includegraphics[width=0.4\textwidth]{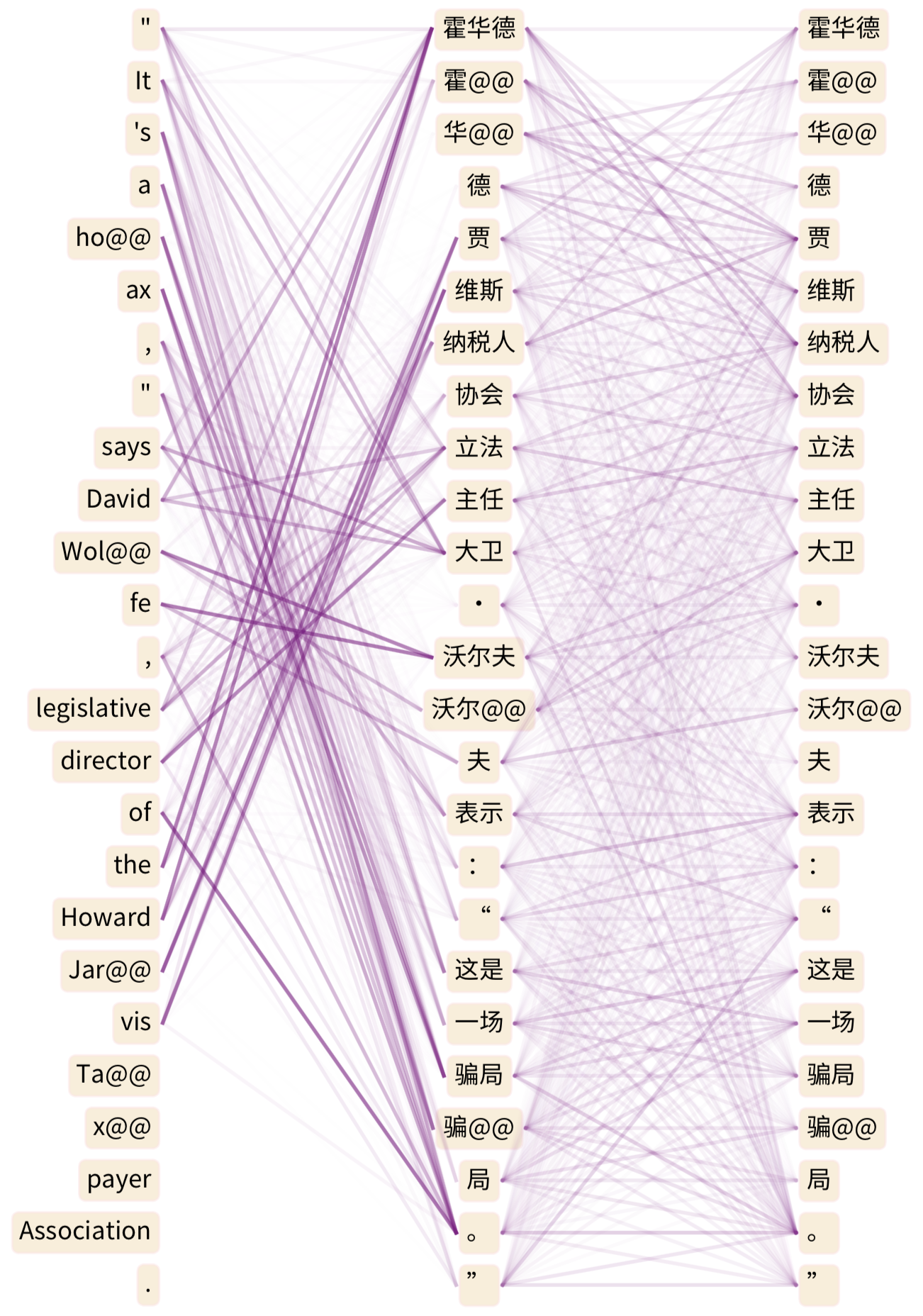}
\includegraphics[width=0.4\textwidth]{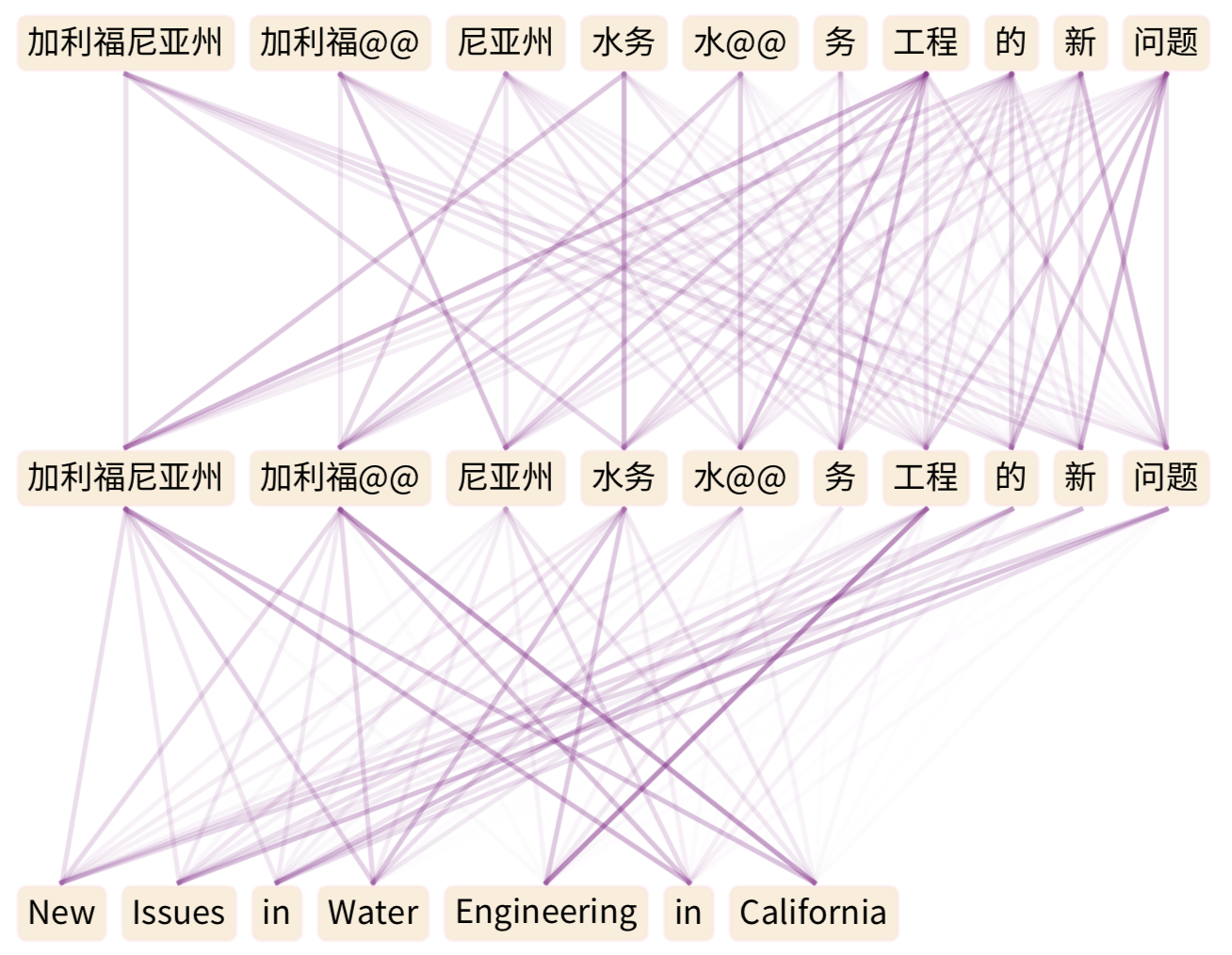}
\includegraphics[width=0.5\textwidth]{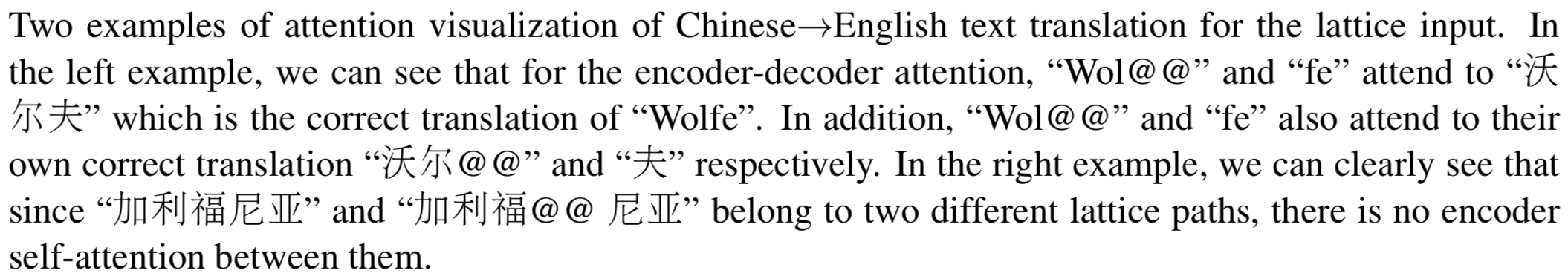}
\caption{Attention visualization for Chinese English translation task.}
\label{fig:zhen_vis}
\end{figure}

\subsection{Chinese English Text Translation}

\begin{table}[h]
\setlength{\tabcolsep}{2pt}
\centering
\resizebox{\columnwidth}{!}{
\begin{tabular}{|l|c|c|}
\hline
Architecture & Inference Inputs & test2017 \\
\hline
Transformer \cite{zhang2018regularizing} & BPE 32K & 23.01 \\
\hline
Transformer-big \cite{hassan2018achieving} & BPE 32K & 24.20 \\
\hline
1. Transformer with BPE 32K & BPE 32K & 24.26 \\
\hline
2. Lattice Transformer from scratch & lattice & 24.71 \\
\hline
3. Lattice Transformer with fine-tuning & lattice & 24.81 \\
\hline
\end{tabular}
}
\caption{BLEU on WMT 2017 Chinese English}
\label{tab:wmt17}
\end{table}

In this experiment, we demonstrate the performance of our lattice transformer when the probability scores are unavailable. 
The comparison baseline method is the vanilla transformer \cite{vaswani2017attention} in both base and big settings.

\subsubsection{Datasets and Settings}
The Chinese to English parallel corpus for WMT 2017 news task contains about 20 million sentences after deduplication. 
For Chinese word segmentation, we use Jieba\footnote{https://github.com/fxsjy/jieba} as the baseline  \cite{zhang2018regularizing,hassan2018achieving}, while the English sentences are tokenized by moses tokenizer. 
Some data filtering tricks have been applied, such as the ratio within $[1/3, 3]$ of lengths between source and target sentence pairs and the count of tokens in both sides ($\leq 200$). 

Then for the Chinese source corpus, we learn the BPE tokenization with 16K / 32K / 48K operations, while for the English target corpus, we only learn the BPE tokenization with 32K operations. 
In this way, each Chinese input can be represented as three different tokenized results, thus being ready to construct a word lattice. 

The hyper-parameters of our model are the same as the setting with the speech translation in previous experiments. 
We follow the optimization convention in \cite{vaswani2017attention} to use ADAM optimizer with Noam invert squared decay. 
All of our lattice transformers are trained on 4 P-100 GPU cards. 
Similar to our comparison method, detokenized cased-sensitive BLEU is reported in our experiment.

\subsubsection{Results}

For our lattice transformer, we have three models trained for comparison. 
First we use the 32K BPE Chinese corpus alone to train our lattice transformer, which is equivalent to the standard transformer with relative positional embeddings. 
Secondly, we train another lattice transformer with the word lattice corpus from scratch. 
In addition, we follow the convention of the speech translation task in previous experiments by fine-tuning the first model with word lattice corpus. 
For each setting, the model evaluated on test 2017 dataset is selected from the best model performed on the dev2017 data. 
The fine-tuning of Lattice Model 3 starts from the best checkpoint of Lattice Model 1.
The BLEU evaluation is shown in Table~\ref{tab:wmt17}, and two examples of attention visualization are shown in Figure~\ref{fig:zhen_vis}. 
Notice that the first two results of transformer-base and -big are directly copied from the relevant references. 
From the result, we can see that our Model 1 can be comparable with the vanilla transformer-big model in a base setting, and significantly better than the transformer-base model. 
We also validate the argument that training from scratch can also achieve a better result than most baselines. 
Empirically, we find an interesting phenomena that training from scratch converges faster than other settings. 

\section{Conclusions}

In this paper, we propose a novel lattice transformer architecture with a controllable lattice attention mechanism that can consume a word lattice and probability scores from the ASR system. 
The proposed approach is naturally applied to both the encoder self-attention and encoder-decoder attention. 
We mainly validate our lattice transformer on speech translation task, and additionally demonstrate its generalization to text translation on the WMT 2017 Chinese-English translation task. 
In general, the lattice transformer can increase the metric BLEU for translation tasks by a significant margin over many baselines.

\section*{Acknowledgments}

We thank Nguyen Bach to provide the script for attention visualization.


\bibliography{acl2019}

\begin{thebibliography}{27}
\expandafter\ifx\csname natexlab\endcsname\relax\def\natexlab#1{#1}\fi

\bibitem[{Adams et~al.(2016)Adams, Neubig, Cohn, and Bird}]{adams2016learning}
Oliver Adams, Graham Neubig, Trevor Cohn, and Steven Bird. 2016.
\newblock Learning a translation model from word lattices.
\newblock In \emph{Interspeech}.

\bibitem[{Ba et~al.(2016)Ba, Kiros, and Hinton}]{ba2016layer}
Jimmy~Lei Ba, Jamie~Ryan Kiros, and Geoffrey~E Hinton. 2016.
\newblock Layer normalization.
\newblock \emph{arXiv preprint arXiv:1607.06450}.

\bibitem[{Bojar et~al.(2018)Bojar, Federmann, Fishel, Graham, Haddow, Koehn,
  and Monz}]{bojar2018findings}
Ond{\v{r}}ej Bojar, Christian Federmann, Mark Fishel, Yvette Graham, Barry
  Haddow, Philipp Koehn, and Christof Monz. 2018.
\newblock Findings of the 2018 conference on machine translation (wmt18).
\newblock In \emph{Proceedings of the Third Conference on Machine Translation:
  Shared Task Papers}, pages 272--303.

\bibitem[{Buckman and Neubig(2018)}]{buckman2018neural}
Jacob Buckman and Graham Neubig. 2018.
\newblock Neural lattice language models.
\newblock \emph{Transactions of the Association for Computational Linguistics},
  6:529--541.

\bibitem[{Devlin et~al.(2018)Devlin, Chang, Lee, and
  Toutanova}]{devlin2018bert}
Jacob Devlin, Ming-Wei Chang, Kenton Lee, and Kristina Toutanova. 2018.
\newblock Bert: Pre-training of deep bidirectional transformers for language
  understanding.
\newblock \emph{arXiv preprint arXiv:1810.04805}.

\bibitem[{Fan et~al.(2018)Fan, Li, Zhou, and Wang}]{fan2018bilingual}
Kai Fan, Bo~Li, Fengming Zhou, and Jiayi Wang. 2018.
\newblock " bilingual expert" can find translation errors.
\newblock \emph{arXiv preprint arXiv:1807.09433}.

\bibitem[{Hassan et~al.(2018)Hassan, Aue, Chen, Chowdhary, Clark, Federmann,
  Huang, Junczys-Dowmunt, Lewis, Li et~al.}]{hassan2018achieving}
Hany Hassan, Anthony Aue, Chang Chen, Vishal Chowdhary, Jonathan Clark,
  Christian Federmann, Xuedong Huang, Marcin Junczys-Dowmunt, William Lewis,
  Mu~Li, et~al. 2018.
\newblock Achieving human parity on automatic chinese to english news
  translation.
\newblock \emph{arXiv preprint arXiv:1803.05567}.

\bibitem[{He et~al.(2016)He, Zhang, Ren, and Sun}]{he2016deep}
Kaiming He, Xiangyu Zhang, Shaoqing Ren, and Jian Sun. 2016.
\newblock Deep residual learning for image recognition.
\newblock In \emph{Proceedings of the IEEE conference on computer vision and
  pattern recognition}, pages 770--778.

\bibitem[{Hochreiter and Schmidhuber(1997)}]{hochreiter1997long}
Sepp Hochreiter and J{\"u}rgen Schmidhuber. 1997.
\newblock Long short-term memory.
\newblock \emph{Neural computation}, 9(8):1735--1780.

\bibitem[{Jan et~al.(2018)Jan, Cattoni, Sebastian, Cettolo, Turchi, and
  Federico}]{jan2018iwslt}
Niehues Jan, Roldano Cattoni, St{\"u}ker Sebastian, Mauro Cettolo, Marco
  Turchi, and Marcello Federico. 2018.
\newblock The iwslt 2018 evaluation campaign.
\newblock In \emph{International Workshop on Spoken Language Translation},
  pages 2--6.

\bibitem[{Khandelwal et~al.(2018)Khandelwal, He, Qi, and
  Jurafsky}]{khandelwal2018sharp}
Urvashi Khandelwal, He~He, Peng Qi, and Dan Jurafsky. 2018.
\newblock Sharp nearby, fuzzy far away: How neural language models use context.
\newblock \emph{arXiv preprint arXiv:1805.04623}.

\bibitem[{Kudo(2018)}]{kudo2018subword}
Taku Kudo. 2018.
\newblock Subword regularization: Improving neural network translation models
  with multiple subword candidates.
\newblock In \emph{Proceedings of the 56th Annual Meeting of the Association
  for Computational Linguistics (Volume 1: Long Papers)}, volume~1, pages
  66--75.

\bibitem[{Lample et~al.(2018)Lample, Ott, Conneau, Denoyer
  et~al.}]{lample2018phrase}
Guillaume Lample, Myle Ott, Alexis Conneau, Ludovic Denoyer, et~al. 2018.
\newblock Phrase-based \& neural unsupervised machine translation.
\newblock In \emph{Proceedings of the 2018 Conference on Empirical Methods in
  Natural Language Processing}, pages 5039--5049.

\bibitem[{Osamura et~al.(2018)Osamura, Kano, Sakti, Sudoh, and
  Nakamura}]{osamura2018using}
Kaho Osamura, Takatomo Kano, Sakriani Sakti, Katsuhito Sudoh, and Satoshi
  Nakamura. 2018.
\newblock Using spoken word posterior features in neural machine translation.
\newblock \emph{architecture}, 21:22.

\bibitem[{Post et~al.(2013)Post, Kumar, Lopez, Karakos, Callison-Burch, and
  Khudanpur}]{post2013improved}
Matt Post, Gaurav Kumar, Adam Lopez, Damianos Karakos, Chris Callison-Burch,
  and Sanjeev Khudanpur. 2013.
\newblock Improved speech-to-text translation with the fisher and callhome
  spanish--english speech translation corpus.
\newblock In \emph{International Workshop on Spoken Language Translation}.

\bibitem[{Rabiner(1989)}]{rabiner1989tutorial}
Lawrence~R Rabiner. 1989.
\newblock A tutorial on hidden markov models and selected applications in
  speech recognition.
\newblock \emph{Proceedings of the IEEE}, 77(2):257--286.

\bibitem[{Sennrich et~al.(2016)Sennrich, Haddow, and
  Birch}]{sennrich2016neural}
Rico Sennrich, Barry Haddow, and Alexandra Birch. 2016.
\newblock Neural machine translation of rare words with subword units.
\newblock In \emph{Proceedings of the 54th Annual Meeting of the Association
  for Computational Linguistics (Volume 1: Long Papers)}, volume~1, pages
  1715--1725.

\bibitem[{Shaw et~al.(2018)Shaw, Uszkoreit, and Vaswani}]{shaw2018self}
Peter Shaw, Jakob Uszkoreit, and Ashish Vaswani. 2018.
\newblock Self-attention with relative position representations.
\newblock In \emph{Proceedings of the 2018 Conference of the North American
  Chapter of the Association for Computational Linguistics: Human Language
  Technologies, Volume 2 (Short Papers)}, volume~2, pages 464--468.

\bibitem[{Sperber et~al.(2017)Sperber, Neubig, Niehues, and
  Waibel}]{sperber2017neural}
Matthias Sperber, Graham Neubig, Jan Niehues, and Alex Waibel. 2017.
\newblock Neural lattice-to-sequence models for uncertain inputs.
\newblock In \emph{Proceedings of the 2017 Conference on Empirical Methods in
  Natural Language Processing}, pages 1380--1389.

\bibitem[{Su et~al.(2017)Su, Tan, Xiong, Ji, Shi, and Liu}]{su2017lattice}
Jinsong Su, Zhixing Tan, Deyi Xiong, Rongrong Ji, Xiaodong Shi, and Yang Liu.
  2017.
\newblock Lattice-based recurrent neural network encoders for neural machine
  translation.
\newblock In \emph{Thirty-First AAAI Conference on Artificial Intelligence}.

\bibitem[{Su et~al.(2018)Su, Fan, Bach, Kuo, and Huang}]{su2018unsupervised}
Yuanhang Su, Kai Fan, Nguyen Bach, C-C~Jay Kuo, and Fei Huang. 2018.
\newblock Unsupervised multi-modal neural machine translation.
\newblock \emph{arXiv preprint arXiv:1811.11365}.

\bibitem[{Sutskever et~al.(2014)Sutskever, Vinyals, and
  Le}]{sutskever2014sequence}
Ilya Sutskever, Oriol Vinyals, and Quoc~V Le. 2014.
\newblock Sequence to sequence learning with neural networks.
\newblock In \emph{Advances in neural information processing systems}, pages
  3104--3112.

\bibitem[{Tai et~al.(2015)Tai, Socher, and Manning}]{tai2015improved}
Kai~Sheng Tai, Richard Socher, and Christopher~D Manning. 2015.
\newblock Improved semantic representations from tree-structured long
  short-term memory networks.
\newblock In \emph{Proceedings of the 53rd Annual Meeting of the Association
  for Computational Linguistics and the 7th International Joint Conference on
  Natural Language Processing (Volume 1: Long Papers)}, volume~1, pages
  1556--1566.

\bibitem[{Vaswani et~al.(2017)Vaswani, Shazeer, Parmar, Uszkoreit, Jones,
  Gomez, Kaiser, and Polosukhin}]{vaswani2017attention}
Ashish Vaswani, Noam Shazeer, Niki Parmar, Jakob Uszkoreit, Llion Jones,
  Aidan~N Gomez, {\L}ukasz Kaiser, and Illia Polosukhin. 2017.
\newblock Attention is all you need.
\newblock In \emph{Advances in Neural Information Processing Systems}, pages
  5998--6008.

\bibitem[{Yu et~al.(2018)Yu, Dohan, Luong, Zhao, Chen, Norouzi, and
  Le}]{yu2018qanet}
Adams~Wei Yu, David Dohan, Minh-Thang Luong, Rui Zhao, Kai Chen, Mohammad
  Norouzi, and Quoc~V Le. 2018.
\newblock Qanet: Combining local convolution with global self-attention for
  reading comprehension.
\newblock \emph{arXiv preprint arXiv:1804.09541}.

\bibitem[{Zhang and Yang(2018)}]{zhang2018chinese}
Yue Zhang and Jie Yang. 2018.
\newblock Chinese ner using lattice lstm.
\newblock In \emph{Proceedings of the 56th Annual Meeting of the Association
  for Computational Linguistics (Volume 1: Long Papers)}, volume~1, pages
  1554--1564.

\bibitem[{Zhang et~al.(2018)Zhang, Wu, Liu, Li, Zhou, and
  Chen}]{zhang2018regularizing}
Zhirui Zhang, Shuangzhi Wu, Shujie Liu, Mu~Li, Ming Zhou, and Enhong Chen.
  2018.
\newblock Regularizing neural machine translation by target-bidirectional
  agreement.
\newblock \emph{arXiv preprint arXiv:1808.04064}.

\end{thebibliography}
\bibliographystyle{acl_natbib}

\end{document}